\documentclass[10pt,twocolumn,letterpaper]{article}

\usepackage{cvpr}
\usepackage{times}
\usepackage{epsfig}
\usepackage{graphicx}
\usepackage{amsmath}
\usepackage{amssymb}
\usepackage{algorithm}
\usepackage[noend]{algpseudocode}
\usepackage{enumitem}

\usepackage[pagebackref=true,breaklinks=true,letterpaper=true,colorlinks,bookmarks=false]{hyperref}

\cvprfinalcopy 

\def\cvprPaperID{****} 

\ifcvprfinal\fi

\newcommand{\dummyfig}[1]{
  \centering
  \fbox{
    \begin{minipage}[c][0.2\textheight][c]{0.45\textwidth}
      \centering{#1}
    \end{minipage}
  }
}
\begin{document}

\title{Egocentric affordance detection with the one-shot \\ geometry-driven Interaction Tensor}

\author{Eduardo Ruiz and Walterio Mayol-Cuevas\\
Department of Computer Science\\
University of Bristol, UK\\
{\tt\small \{er13827,cswwmc\}@bristol.ac.uk}
}

\maketitle

\begin{abstract}
In this abstract we describe recent \cite{ruiz2018,ruiz:2018b} and latest work on the determination of affordances in visually perceived 3D scenes. Our method builds on the hypothesis that geometry on its own provides enough information to enable the detection of significant interaction possibilities in the environment. The motivation behind this is that geometric information is intimately related to the physical interactions afforded by objects in the world. The approach uses a generic representation for the interaction between everyday objects such as a mug or an umbrella with the environment, and also for more complex affordances such as humans \textit{Sitting} or \textit{Riding} a motorcycle. Experiments with synthetic and real RGB-D scenes show that the representation enables the prediction of affordance candidate locations in novel environments at fast rates and from a single (one-shot) training example. The determination of affordances is a crucial step towards systems that need to perceive and interact with their surroundings. We here illustrate output on two cases for a simulated robot and for an Augmented Reality setting, both perceiving in an egocentric manner. 
\end{abstract}
\section{Introduction}

\begin{figure}[t]
        \centering
        \IfFileExists{scannet_examples-eps-converted-to.pdf}{\includegraphics[width=.4\textwidth]{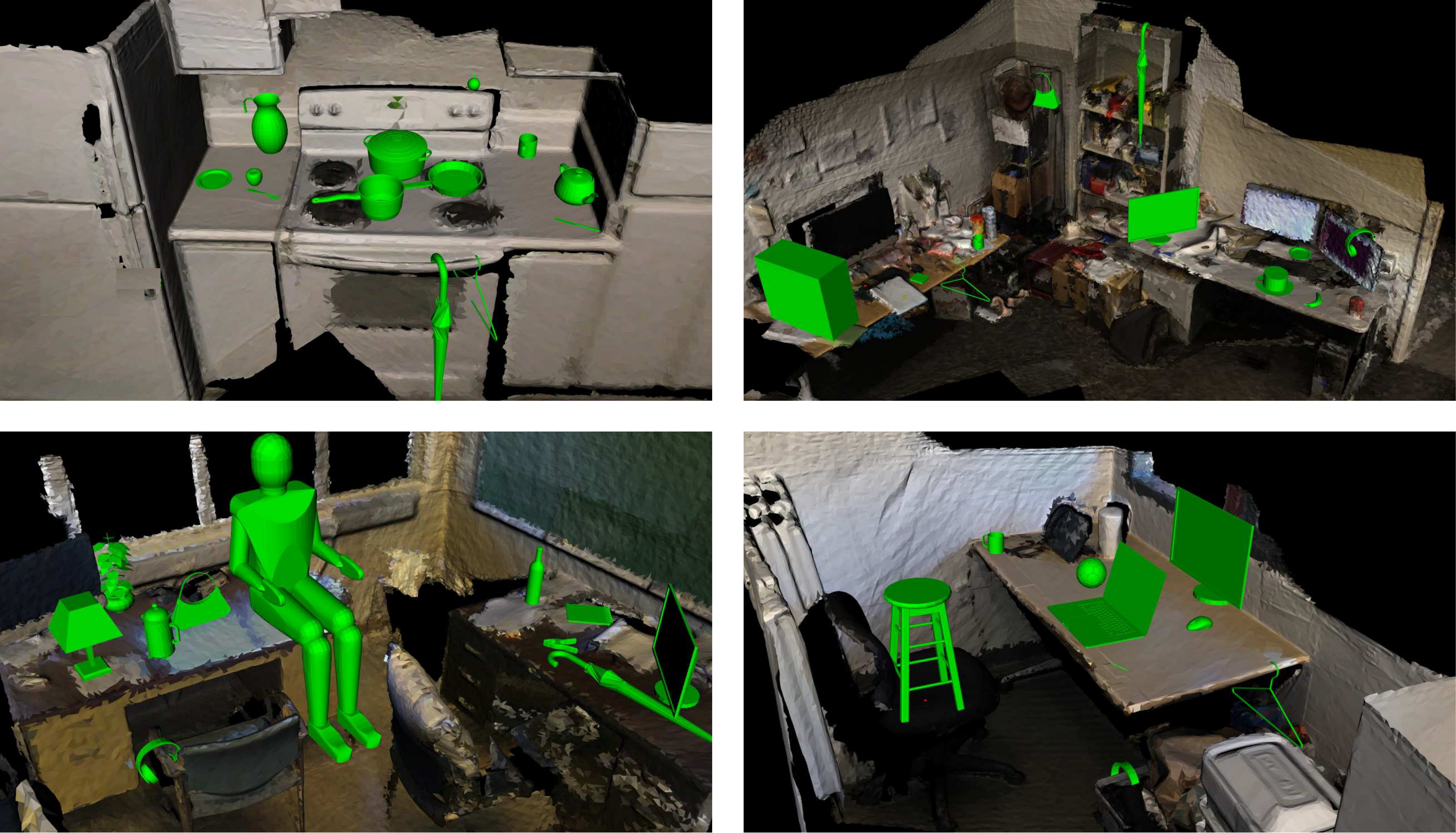}}{\dummyfig{toyInteracionExample}}
      \caption{Affordance detections (green objects and their poses) determined by our geometry-based method. We are able to determine over 80 affordances in  real-time on never before seen RGB-D scenes.}
        \label{fig: examples first}
\vspace{-2mm}
\end{figure}

Agents that need to act on their surroundings can significantly benefit from the perception of their interaction possibilities or affordances. The concept of affordance is innately interlinked and founded for egocentric perception, and the term coined by James J. Gibson \cite{gibson1977} within the field of ecological perception. For Gibson, affordances are action opportunities in the environment that are \textit{directly} perceived by the observer. According to this, the goal of vision is to recognise the affordances rather than the elements or objects in the scene. The concept of affordances calls for an approach to visual perception that is free from non-action representations, and that is there to help the agent to interact with the world.
Following Gibson's call for a {\it direct} perception of affordances, and motivated by studies in neuroscience showing that affordance detection does not require semantic reasoning \cite{VINGERHOETS2009conceptual}; we
hypothesise that geometric information or shape provides enough information for an agent to directly perceive the interaction opportunities in its surroundings. Examples of our geometry-driven affordance detection approach are shown in Fig. \ref{fig: examples first}. As detailed later on, the detection is agnostic to semantics and complex representations of the input scene.

\textbf{Related work} \enskip
Much of the attention given to the problem of affordances has focused on the classification of object instances in the world\cite{Do2018}, internal symbolic relationships \cite{Zhu2014} or semantic category information\cite{Chuang2018}, which strongly undermines the idea of \textit{direct} and economical perception of affordances
proposed by Gibson. But to {\it directly} determine affordances has faced many dilemmas, namely the challenging problems of visually recovering the relevant properties of the environment in a robust and accurate manner.  

We argue that in order to truly perceive affordances in a way that is most useful for agents, there is a need for methods that are agnostic to object categories and free from complex feature representations; methods of a generic nature that allow for the simple yet robust description of multiple affordances.  We hypothesise that geometry on its own provides enough information to robustly and generically characterise affordances.

\section{Our approach}
We concentrate on the subclass of affordances between rigid objects. Affordances such as “where can I hang this?”, place this, ride, fill, and similar. We do this by specifying a geometry-driven interaction tensor that aims to capture the way in which the affordance manifests between a pair of objects. In contrast with previous approaches, our algorithm is able to generalise from a single training example to completely novel environments, i.e. one-shot learning. Here we describe the core of our approach, namely the affordance representation (Interaction Tensor) and the algorithm that allows for fast one-shot detections.

\begin{figure*}[!ht]
        \centering
        \IfFileExists{tensor_testing-eps-converted-to.pdf}{\includegraphics[width=0.85\textwidth]{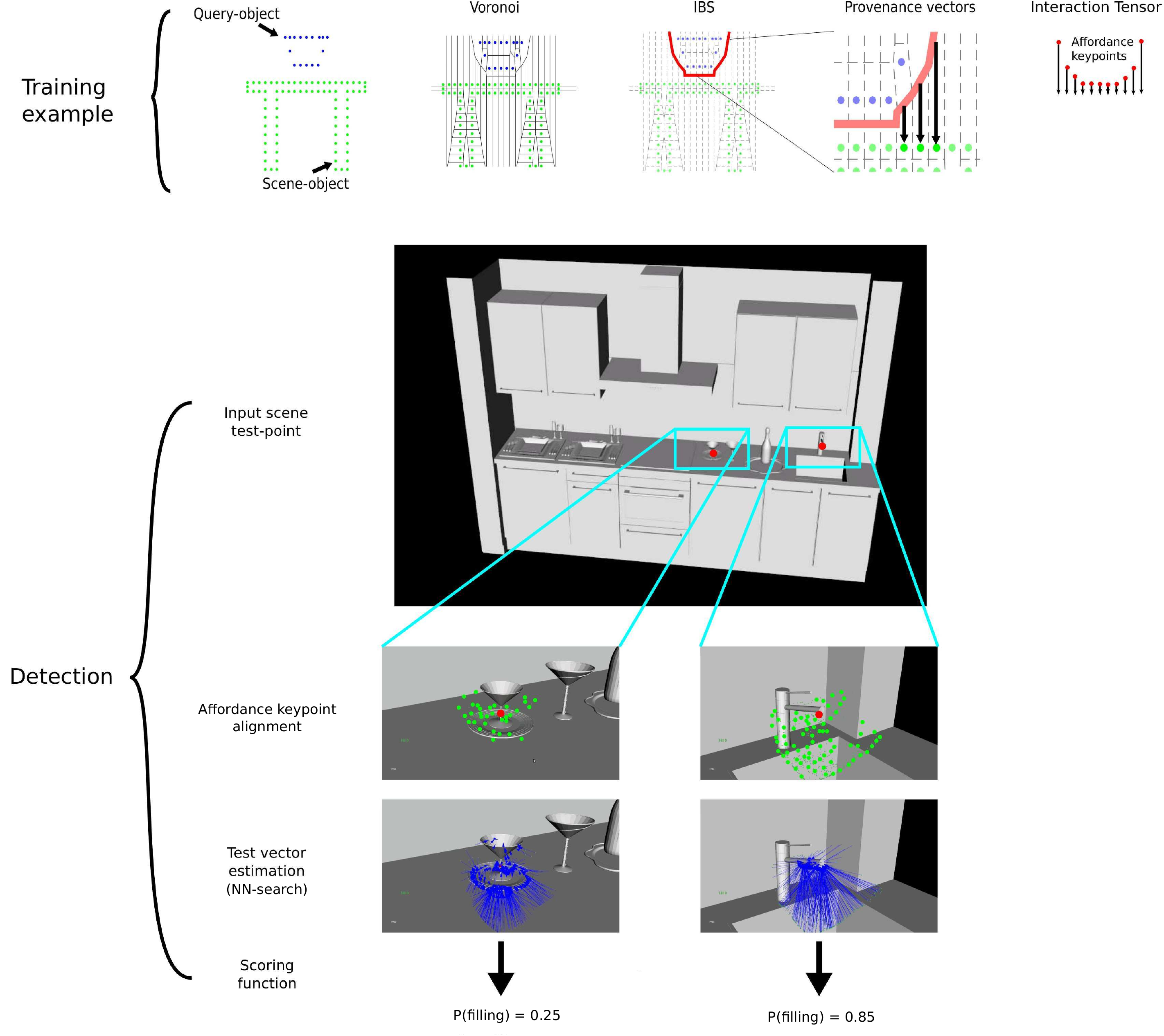}}{\dummyfig{toyInteracionExample}}
      \caption{Top row (Training 2D example) illustrates the computation of an interaction tensor for an affordance of interest, in this particular case \textit{Placing} a bowl. Figure on the bottom (Detection) shows the approach followed to perform affordance detection on novel scenarios. For clarity purposes the scene shown is a synthetic kitchen; however, the detection algorithm allows to follow the same approach for RGB-D scenes.}
        \label{fig: approach}
\end{figure*}

\textbf{The Interaction Tensor} \enskip
The Interaction Tensor ($\mathbf{iT}$) \cite{ruiz2018} is a vector field representation able to characterise the static interactions between 2 generic entities (e.g. objects) in 3D space. This proposed representation builds on the Interaction Bisector Surface (IBS)\cite{Zhao2014} concept and extends its robustness by three main factors:
\begin{enumerate}[topsep=4pt,itemsep=2pt]
 \item Proposing a representation suitable for visually generated data, e.g. pointclouds
 \item Increases robustness by encoding the locations in the interacting entities that contributed to the computation of their bisector  ---provenance vectors
 \item Introduces a straight forward descriptor that allows for real-time prediction of affordance candidate locations on RGB-D data ---affordance keypoints
\end{enumerate}
Using direct, sparse sampling over the $\mathbf{iT}$ allows for the determination of geometrically similar interactions from a single \textit{training} example; this sampling comprises what we call {\it affordance keypoints}, which serve to more quickly judge the likelihood of an affordance at a test point in a scene. The $\mathbf{iT}$ is straightforward to compute and tolerates well changes in geometry, which provides a good generalisation to unseen scenes from a single example. The key steps include an example affordance from a simulated interaction, the computation of the IBS between an object (query-object) and scene (or scene-object), and estimating provenance vectors which are the vectors used in the computation of points on the bisector surface. Top row (Training example) in Fig. \ref{fig: approach} shows the elements and the process involved in computing an affordance $\mathbf{iT}$ for \textit{Placing} a bowl.

\textbf{Fast one-shot affordance detection} \enskip 
In order to make fast affordance detection in a novel scenario without computing the full $\mathbf{iT}$ descriptor, we perform an approximation of the descriptor via a Nearest Neighbour (NN) search. This can be done by taking advantage of the \textit{provenance vectors} from the training example; these vectors account for regions in the scene that contributed to the computation of the bisector surface. The proposed algorithm uses this information
to investigate whether those regions exist in a novel scenario, these regions would allow computing the same or a similar $\mathbf{iT}$. In this sense, the NN-search is used to investigate if the point in the scene required to compute a point on the $\mathbf{iT}$ exists; or more precisely, if the point in the scene is where is expected to be. The detection pipeline is illustrated in the bottom diagram (Detection) of Fig. \ref{fig: approach}. Full description of our methods is available at \cite{ruiz2018} and \cite{ruiz:2018b}.
\begin{figure*}[!h]
        \centering
        \IfFileExists{augmented-eps-converted-to.pdf}{\includegraphics[width=\textwidth]{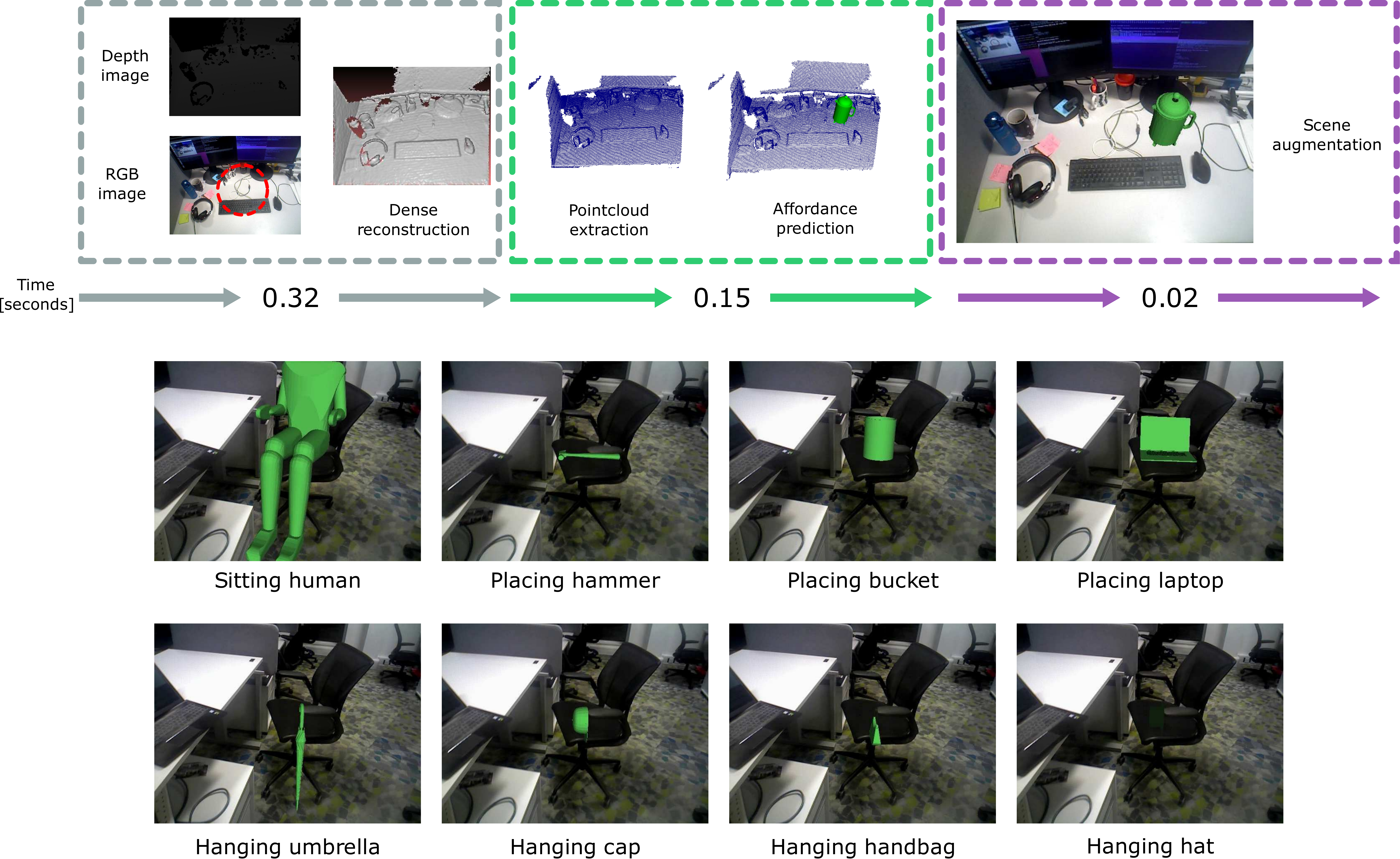}}{\dummyfig{toyInteracionExample}}
      \caption{Top row shows a demonstration for Augmented Reality. The input to the system is the RGB and depth images of the current scene. The red circle in the RGB image on the left illustrates the ROI used for detections. Bottom rows show example predictions for \textit{Sitting, Placing} and \textit{Hanging} objects in an office environment.}
        \label{fig: lantern}
\vspace{-2mm}
\end{figure*}
\begin{figure*}[!h]
        \centering
        \IfFileExists{sim_robot-eps-converted-to.pdf}{\includegraphics[width=0.8\textwidth]{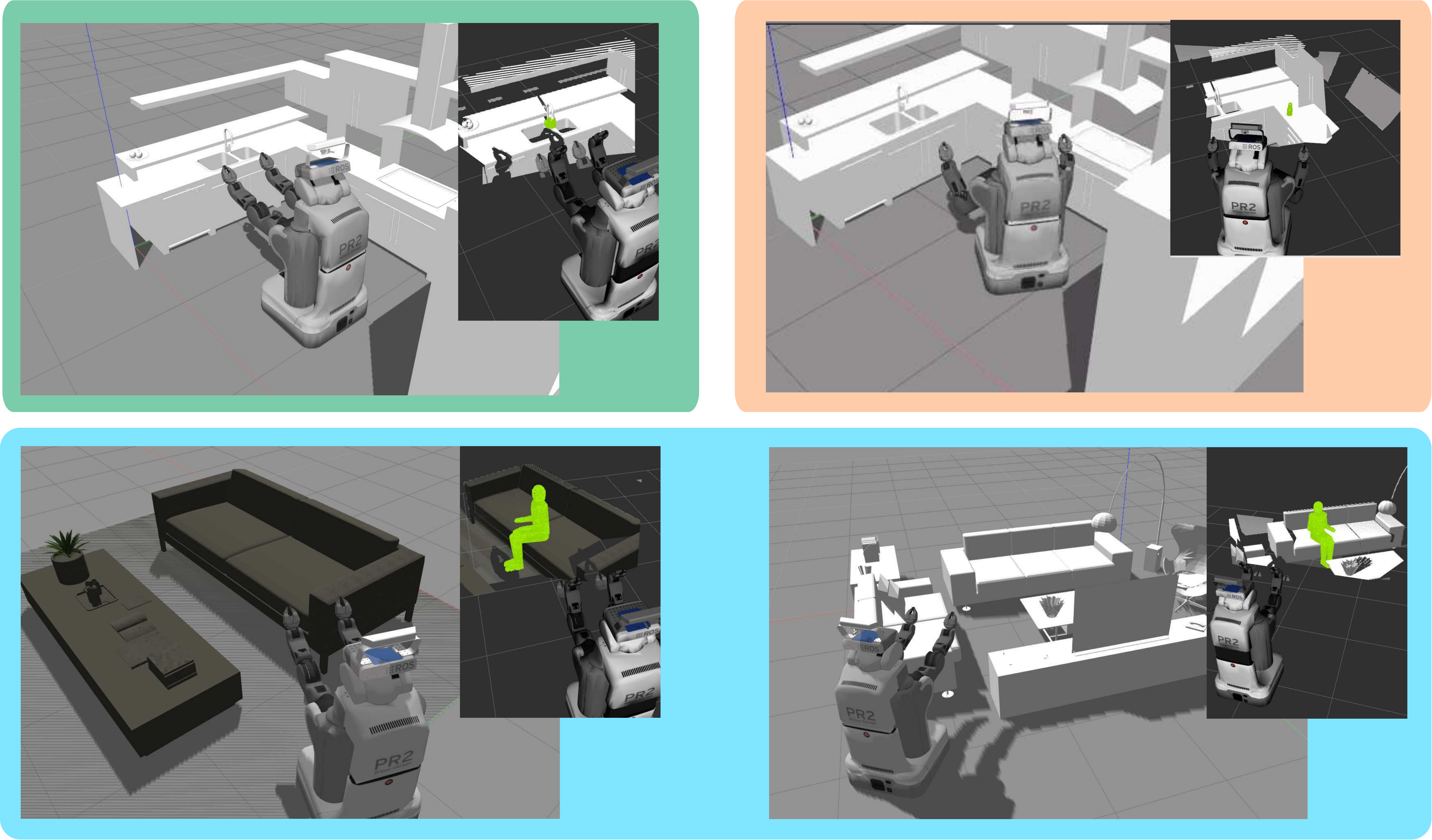}}{\dummyfig{toyInteracionExample}}
      \caption{The fast rates at which detections are made allows the integration of the affordance detection algorithm into the perception pipeline of a robotic system. For this simulation, the input to the detection algorithm is directly the pointcloud as captured by the RGB-D sensor. Top row images show a robot in a kitchen environment where \textit{Filling} (left) and \textit{Placing}(right) affordances have been detected. Images on the bottom row show \textit{Sitting} affordance detections performed while the robot navigates in a living-room. In all images query-objects are shown in green.}
        \label{fig: robot}
\vspace{-2mm}
\end{figure*}
\section{For Robots and AR}
We leverage a state-of-the-art and publicly available dense mapping system \cite{newcombe2011} paired with an RGB-D sensor to recover a 3D representation of the scene in front of the camera. Our implementation of the affordance detection algorithm leverages the parallelisation capabilities of commodity desktop hardware. As a reference, our algorithm running in a PC with a Titan X GPU allows for the simultaneous detection of up to 84 affordances at 10 point locations of the input scene in under 1 second. Fig. \ref{fig: lantern} summarises the computation times involved in the current implementation.

Our experiments include multiple affordances of everyday objects such as cups, mugs, bowls, etc. and also detections of \textit{human} affordances such as \textit{Sitting} or \textit{Riding}. Here we show qualitative results of the algorithm of its application in robotics systems (Fig. \ref{fig: robot}) as well as for augmented reality (Fig. \ref{fig: lantern}). The scenes used for our experiments include publicly available data such as \cite{scannet2017}, amongst others of our own. Code and data of our core affordance detection approach have been made publicly available\footnote{https://github.com/eduard626/interaction-tensor}.

\section{Conclusion}
We have developed a tensor field representation to characterise the interactions between pairs of objects. This representation and the proposed algorithm allow for real-time and multiple affordance detections in novel environments \textit{training} from a single example. In this abstract we showed results of the application of the proposed approach for robotic perception and scene augmentation in mixed reality systems. Overall, we see this work as an effort to motivate further advancing of approaches in Vision that are more ecological in nature and consider the relationship between the scene and the perceiving agent.
{\small
\bibliographystyle{ieee_fullname}
\bibliography{egbib}
}

\end{document}